\documentclass[10pt,twocolumn,letterpaper]{article}

\usepackage{iccv}
\usepackage{times}
\usepackage{epsfig}
\usepackage{graphicx}
\usepackage{amsmath}
\usepackage{amssymb}
\usepackage{booktabs}
\usepackage{multirow}
\usepackage{algorithm}
\usepackage{algpseudocode}
\newcommand{\mypm}{\,$\pm$\,}
\newcommand{\minisection}[1]{\vspace{0.02in} \noindent {\bf #1}\ }
\usepackage[breaklinks=true,bookmarks=false]{hyperref}
\def\ourmethod{{\emph{ScrollNet}}\xspace} 

\iccvfinalcopy

\ificcvfinal\fi

\begin{document}

\title{ScrollNet: Dynamic Weight Importance for Continual Learning}

\author{Fei Yang$^{1}$, Kai Wang$^{1}$, Joost van de Weijer$^{1}$\\
$^{1}$ Computer Vision Center, Universitat Autònoma de Barcelona, Barcelona, Spain\\
{\tt\small \{fyang,kwang,joost\}@cvc.uab.es}
}

\maketitle
\ificcvfinal\thispagestyle{empty}\fi

\begin{abstract}
The principle underlying most existing continual learning (CL) methods is to prioritize stability by penalizing changes in parameters crucial to old tasks, while allowing for plasticity in other parameters. The importance of weights for each task can be determined either explicitly through learning a task-specific mask during training (e.g., parameter isolation-based approaches) or implicitly by introducing a regularization term (e.g., regularization-based approaches).
However, all these methods assume that the importance of weights for each task is unknown prior to data exposure. In this paper, we propose ScrollNet as a scrolling neural network for continual learning. ScrollNet can be seen as a dynamic network that assigns the ranking of weight importance for each task before data exposure, thus achieving a more favorable stability-plasticity tradeoff during sequential task learning by reassigning this ranking for different tasks. Additionally, we demonstrate that ScrollNet can be combined with various CL methods, including regularization-based and replay-based approaches. Experimental results on CIFAR100 and TinyImagenet datasets show the effectiveness of our proposed method. We release our code at \url{https://github.com/FireFYF/ScrollNet.git}.
\end{abstract}

\section{Introduction}
\label{sec:intro}
\begin{figure}[t]
  \centering
   \includegraphics[width=0.99\linewidth]{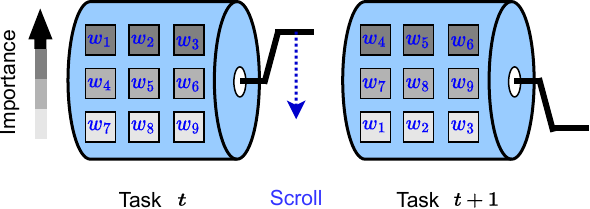}
   \caption{
   Our proposed method employs a dynamic network to pre-assign weight importance before training. This dynamic network ensures that the most crucial weights for a given task $t$ are consistently assigned at the top of the `roller' (a metaphor for a neural network). Once the current task is completed, we scroll the roller to readjust the order of weight importance for the subsequent task. This process aims to strike a balance between stability and plasticity, allowing the model to retain previously learned knowledge while remaining adaptable to new tasks.
   }
   \label{fig:scroll}
   \vspace{-0.2cm}
\end{figure}

In human life, knowledge is consistently acquired and accumulated. However, deep learning models often experience knowledge forgetting, a phenomenon known as catastrophic forgetting~\cite{kirkpatrick2017overcoming,mccloskey1989catastrophic}, when they are exposed to a series of tasks. 
To address this issue, continual learning (CL)~\cite{parisi2019continual,delange2021continual,masana2022class}, also known as lifelong learning, has emerged as a vital research direction for a variety of learning and representation tasks (e.g., image classification\cite{serra2018overcoming, masana2022class}, semantic segmentation~\cite{douillard2021plop,yu2022self}, generative models~\cite{wu2018memory}, object detection, autonomous driving~\cite{verwimp2023clad}). Continual learning aims to prevent the loss of previously acquired knowledge in neural networks over time.

This issue is intricately connected to the stability-plasticity dilemma~\cite{kong2022balancing,mirzadeh2020understanding}. More precisely, when learning sequentially, the network needs to possess the ability to incorporate new knowledge (plasticity) while also maintaining stability to prevent forgetting previously learned tasks. Nevertheless, the stability-plasticity dilemma suggests that achieving both high plasticity and high stability simultaneously is challenging. Various continual learning approaches have been proposed to address this dilemma, which can be broadly categorized as follows: \textbf{regularization-based methods}~\cite{kirkpatrick2017overcoming,aljundi2018memory}
add a regularization term to the objective function which impedes changes to the parameters deemed relevant to previous tasks; \textbf{replay-based methods}~\cite{rebuffi2017icarl,lopez2017gradient} prevent forgetting by including data from previous tasks, stored either in an episodic memory or via a generative model; \textbf{parameter isolation-based methods}~\cite{serra2018overcoming,rusu2016progressive} propose to modify the update rule of parameters of neural network to minimize the inter-task interference. 

The key to obtaining a balanced stability-plasticity tradeoff is to acquire the importance of weights for each task. This information can then be used to prioritize stability by penalizing changes in parameters crucial to previous tasks while allowing for plasticity in other parameters. The existing CL methods mentioned above determine the importance of weights for different tasks in two different ways: explicitly or implicitly. The explicit approach, such as parameter isolation-based methods~\cite{mallya2018packnet, mallya2018piggyback, serra2018overcoming}, learns a task-specific mask on parameters (or neurons) by introducing a sparsity loss. On the other hand, the implicit approach, like replay-based~\cite{riemer2018learning,tang2021layerwise,liu2020generative} and regularization-based~\cite{li2017learning, kirkpatrick2017overcoming, aljundi2018memory} methods, utilizes memorized old data or regularization terms on objective functions to acquire the importance of weights. However, both approaches assume that the importance of weights for each task is unknown prior to data exposure.

In contrast to these approaches, we propose a novel method to pre-assign the weight importance for each task. Our inspiration comes from an easily overlooked characteristic of dynamic networks~\cite{han2021dynamic, yang2021slimmable, huang2017multi}. Dynamic networks were originally proposed for efficiency, enabling them to adapt their structures or parameters during inference. For example, they can allocate computations on demand by selectively activating model components (e.g., layers~\cite{wolczyk2021zero,kaya2019shallow,bolukbasi2017adaptive}, channels~\cite{yu2018slimmable,jacobs1991adaptive,eigen2013learning}, or sub-networks~\cite{liu2018dynamic,frosst2017distilling,kontschieder2015deep}) while avoiding performance degradation. The hidden characteristic under this property of dynamic networks is weight importance ranking. By splitting the network into small modules with different parameters, weight importance can be assigned by defining the composition of the sub-network in one dynamic network. For instance, the parameters involved in the smallest sub-network will be the most important ones for the current task. Note that sub-networks of different sizes can be mutually inclusive. This allows us to assign different importance to all parameters in the network even before touching training data. The final realization of this pre-assignment of weight importance is achieved through the optimization of the whole dynamic network.

Once we set the weight importance for one task, the next problem is how to use this information to achieve a balance between stability and plasticity in the continual learning of sequential tasks. Following the principle mentioned above, we need to re-order the weight importance before starting the next task. As shown in Figure~\ref{fig:scroll}, we propose a ``scrolling'' operation applied to the modules, which will assign the least importance to the parameters for the next task, which are the most important ones for the current task, and vice versa. Thus, we realize dynamic weight importance in sequential tasks. We name this novel and simple method \emph{Scrollnet}. To the best of our knowledge, this is the first work that explores weight importance pre-assignment prior to data exposure in a continual learning setting. Additionally, we emphasize that \ourmethod is orthogonal to various CL approaches, such as regularization-based and replay-based methods.

In summary, the main contributions of this work are as follows:

\begin{itemize}
    \item We propose \ourmethod, a novel continual learning method that can pre-assign the weight importance before starting a new task. Along with the proposed ``Scrolling'' strategy, ScrollNet can achieve dynamic weight importance and explicitly strike a balance between stability and plasticity.
    \item Our method is orthogonal to various continual learning approaches, and it improves performance when combined with them.
    \item We conduct extensive experiments on CIFAR100 and TinyImageNet. The experimental results demonstrate the effectiveness of our proposed method.
\end{itemize}

\section{Related work}
\label{sec:relatedwork}

\subsection{Continual learning}
\label{sec:relatedwork_cl}
Continual learning~\cite{de2019continual,masana2022class} methods can be loosely categorized into three groups of approaches: regularization-based, replay-based and
parameter-isolation methods. We provide a brief overview of each approach below.

\minisection{Regularization-based methods.} The majority of these approaches add a regularization term to the loss function which impedes changes to the parameters deemed relevant to previous tasks. The difference depends on how to estimate relevance, and these methods can be further divided into data-focused~\cite{li2017learning} and prior-focused~\cite{kirkpatrick2017overcoming,buzzega2020dark,aljundi2018memory}. Data-focused methods use knowledge distillation from previously-learned models~\cite{liu2020generative,wu2018memory,li2017learning}. Prior-focused methods~\cite{kirkpatrick2017overcoming,zenke2017continual} estimate the importance of model parameters as a prior for the new model.
In recent times, there have been several notable studies that specifically concentrate on enforcing weight updates that lie within the null space of the feature covariance~\cite{wang2021training,tang2021layerwise}.
In this paper, we apply our method over several regularization-based methods (including EWC~\cite{kirkpatrick2017overcoming}, MAS~\cite{aljundi2018memory} and LwF~\cite{li2017learning}) to verify our methods.

\minisection{Replay-based methods.} These approaches usually use memory and replay/rehearsal mechanism to recall a small episodic memory of previous tasks while training new tasks thus reduce the loss in the previous tasks. There are two main strategies: exemplar replay~\cite{buzzega2020dark,riemer2018learning,tang2021layerwise,chaudhry2019tiny,chaudhry2018efficient} and pseudo-replay~\cite{shin2017continual,wu2018memory,hayes2020remind,liu2020generative}. The former stores a few training samples (called exemplars) from previous tasks. The latter uses generative models learned from previous data distributions to synthesize data. 
In this paper, we also verify the effectiveness of our method combined with several exemplar replay-based methods, namely iCaRL~\cite{rebuffi2017icarl}, BiC~\cite{wu2019large}, LUCIR~\cite{hou2019learning}, etc.

\minisection{Parameter isolation-based methods.}
This branch tries to learn a sub-network for each task in a shared network~\cite{mehta2021continual}. 
In particular, Piggyback/PackNet~\cite{mallya2018piggyback,mallya2018packnet} iteratively assigns parameter subsets to consecutive tasks by constituting binary masks. 
SupSup~\cite{wortsman2020supermasks} also finds masks in order to assign different subsets of the weights for different tasks. 
HAT~\cite{serra2018overcoming} incorporates task-specific embeddings for attention masking. 
Progressive Neural Network~\cite{rusu2016progressive} allocates a sub-network for each task in advance and progressively concatenates previous sub-networks while freezing parameters allocated to previous tasks. \cite{von2019continual} also proposes task-conditional hypernetworks for continual learning. \cite{masse2018alleviating} proposes nonoverlapping sets of units being active for each task. 
CCGN~\cite{abati2020conditional} proposes task-specific convolutional filter selection for continual learning, but they require a significantly large number of parameters to represent the masks for each task. 
This type of methods is also developed for the case where no forgetting is allowed in TFM~\cite{masana2020ternary}.
In general, this branch is always restricted to the task-aware (task incremental) setting. Thus, they are more suitable for learning a long sequence of tasks when a task oracle is present. 

Similar to parameter isolation-based approaches, our proposed method also segments the entire network into distinct sub-networks by employing a dynamic neural network architecture. However, in contrast to parameter isolation methods, our approach entails the direct and manual specification of ``masks'' for sub-networks, thereby explicitly assigning weight importance to each task even before touching the training data.
Moreover, our method extends its applicability to the more general class-incremental learning setup. Importantly, our approach remains independent of regularization-based and replay-based approaches, establishing its orthogonality to those methods.

\subsection{Dynamic networks}
\label{sec:relatedwork_dn}
Different from the ``dynamic networks (with sub-networks)'' mentioned in Section.~\ref{sec:relatedwork_cl}, which are specifically designed in various parameter isolation-based methods for overcoming catastrophic forgetting in continual learning, the related works we will briefly review here are about those that were originally proposed for inference efficiency. As opposed to static networks, dynamic networks can modify their structure or parameters and control the computation cost during inference in advance. Depending on the different types of ``dynamic'' in the architecture, the dynamic network methods can be grouped as dynamic depth, dynamic width, and dynamic routing.

\minisection{Dynamic depth.}
The architecture with dynamic depth can reduce redundant computation by performing inference with variable depth. The realization approaches include \textit{early exiting}~\cite{teerapittayanon2016branchynet, bolukbasi2017adaptive, huang2017multi, kaya2019shallow, wolczyk2021zero}, which involves executing only shallow layers for ``easy'' samples and avoiding the expensive computation cost of full layers, or layer skipping~\cite{graves2016adaptive, wang2018skipnet, veit2018convolutional}, which selectively skips intermediate layers depending on the complexity of input samples.

\minisection{Dynamic width.}
By performing inference with variable \textit{width} and comparing it with dynamic \textit{depth}, the dynamic \textit{width} architecture exhibits finer-grained control over computation costs. Although all layers are executed, multiple units (e.g., channels, neurons, or branches) within those layers can be selectively activated. Various implementations of the dynamic width architecture have been proposed, such as skipping neurons in fully-connected layers~\cite{bengio2013estimating, cho2014exponentially}, skipping branches in mixture-of-experts~\cite{jacobs1991adaptive,eigen2013learning}, and skipping channels in CNNs~\cite{huang2017multi,yu2018slimmable}. In this paper, we build our proposed \ourmethod upon \cite{yu2018slimmable}. 

\minisection{Dynamic routing.}
The aforementioned approaches adapt the computation cost by adjusting the \textit{depth} or \textit{width} of the architectures. An alternative direction involves creating diverse network forms with multiple potential inference paths and conducting dynamic routing within these networks to adjust the computational graph for different samples. These related methods include path selection in multi-branch structures~\cite{odena2017changing, liu2018dynamic} and tree-structured networks~\cite{kontschieder2015deep, frosst2017distilling}.
\section{Continual learning with a scrolling neural network}
In this section, we describe our proposed dynamic network-based continual learning (CL) method, called \emph{ScrollNet}.
\emph{ScrollNet} can directly assign the weight importance for each task with a dynamic neural network. After training for each task, the model will ``scroll'' the parameter importance assignments before training for the next task, which means reassigning the ranking of weight importance for different tasks. We emphasize that the proposed \emph{ScrollNet} is orthogonal to regularization-based and replay-based CL approaches, which means it can be combined with those methods to achieve better performance.

\noindent \textbf{Problem Statement.}
Consider a supervised continual learning scenario, a learner needs to solve $T$ tasks sequentially without catastrophic forgetting of old tasks. We denote that $\mathcal{D}_t=\{\mathcal{X}_{t}, \mathcal{Y}_{t}\}$ is the dataset of task $t$, composed of a set of input images $\mathcal{X}_{t}$ and corresponding labels $\mathcal{Y}_{t}$. We assume a neural network $f(\cdot;{\theta})$, parameterized by the model weights ${\theta}$ and a standard continual learning scenario aims to learn a sequence of tasks by minimizing the optimization problem at each step $t$: 
\begin{equation}
{\arg \min}_{{\theta}} \mathcal{L}\left(f(\mathcal{X}_{t}; \theta), \mathcal{Y}_{t}\right),
\label{cl_loss}
\end{equation}
where $\mathcal{L}(\cdot, \cdot)$ is a cross-entropy loss for image classification. $\mathcal{D}_t$ for task $t$ is only accessible when learning task $t$. Note that replay-based continual learning approaches allow memorizing a small portion of data from old tasks.

To assign the weight importance for each task, we firstly split the weights $\theta$ into $N$ non-overlapped sets: $\theta = \{\theta_{w_{1}}, \theta_{w_{2}}, ..., \theta_{w_{N}}\}$. We assume the list of these weight sets has a descending order of importance at the task $t$, then the optimization problem during training will be
\begin{equation}
\begin{split}
{\arg \min}_{\theta}&\mathcal{L}\left(f(\mathcal{X}_{t}; \{\theta_{w_{1}}, \theta_{w_{2}}, ..., \theta_{w_{N}} \}), \mathcal{Y}_{t}\right),
\\
&\text{subject to~} I(\theta_{w_{1}}) > I(\theta_{w_{2}}) > .... > I(\theta_{w_{N}}),
\end{split}
\label{cl_loss_rank}
\end{equation}
where $I(\cdot)$ stands for the importance score. In the next sections, we describe how to realize this assignment of weights ranking and how to reassign it in sequential tasks.
\begin{figure*}[t]
  \centering
   \includegraphics[width=1\linewidth]{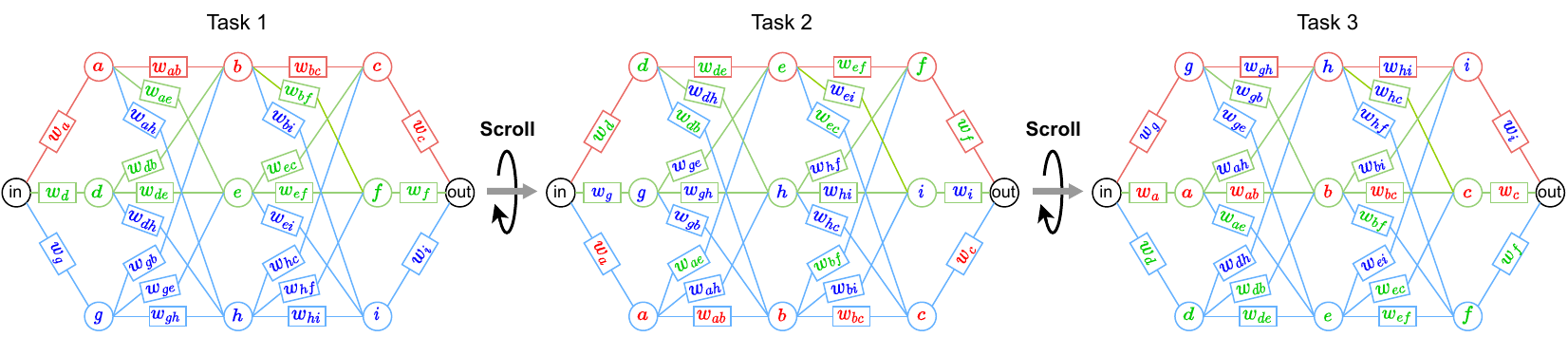}
   \vspace{-0.3cm}
   \caption{
   An illustration figure for our proposed \emph{ScrollNet}. We use color-coded connections to represent the importance of parameters for each task.
   The red connections indicate the most important parameters, the green connections represent the second most important parameters, and the blue connections denote the third most important parameters.
   Before starting a new task, a ``scroll" operation is performed, which reorders the parameter importance. This reassignment ensures that the upcoming task's most crucial parameters are appropriately positioned at the top of the network.
   }
   \label{ScrollNet}
\end{figure*}

\subsection{Weight importance assignment via a dynamic network}
Dynamic networks aim to adapt their structures or parameters to the input during inference, therefore enjoy favorable properties which are absent in static models, such as efficiency, representation power, adaptiveness, etc. Rather than pursuing these characteristics of the dynamic network, here we leverage it to assign the ranking of weight importance. Specifically, we utilize a slimmable neural network~\cite{yu2018slimmable}, which comprises multiple sub-networks within $f(\cdot; \theta)$. To assign the desired weight ranking at the first task as shown in Eq.~\ref{cl_loss_rank}, we can predefine $N$ sub-networks before training as
\begin{equation}
f_{1}(\cdot; \theta^{1} = \theta_{w_{1}}), f_{2}(\cdot; \theta^{2} = \theta_{w_{1}} \cup \theta_{w_{2}}), ..., f_{N}(\cdot; \theta^{N} = \theta).
\label{ranking_1}
\end{equation}
Finally, we can realize this pre-assigned ranking by using the following loss during training:
\begin{equation}
\mathcal{L}_{dynamic} = \sum_{n=1}^N \mathcal{L}\left(f_{n}(\mathcal{X}_{t}; \theta^{n}), \mathcal{Y}_{t}\right).
\label{loss_dynamic}
\end{equation}
As we can observe from this loss function, $\theta^{1}$ serves as the foundational sub-network for task $t$ and holds the highest significance. On the other hand, $\theta^{2}$ and others, despite carrying more parameters, primarily serve to enhance the performance of this foundation, indicating a comparatively lesser importance.
An illustration is presented in Figure~\ref{ScrollNet} (see Task 1), where red connections represent one of the sub-networks with the most significant parameters, while the combination of red and green connections represents another sub-network with the first and second most important parameters.
\subsection{Scrolling neural network for continual learning}
In the previous section, we addressed the assignment and realization of weight importance ranking before and during training for each task. The challenge now lies in leveraging this knowledge to achieve a better stability-plasticity tradeoff in continual learning. As mentioned earlier, the principle used for overcoming catastrophic forgetting is penalizing changes in parameters that are important for previous tasks while allowing updates to less important parameters. Building upon this principle, we propose a parameter ``scrolling'' approach before starting the current task. As shown in Figure~\ref{ScrollNet}, we always assign the most important parameters on the top/red connections (corresponding to the sub-network $\theta_{1}$ in Eq.~\ref{loss_dynamic}) and less important parameters on the bottom connections (i.e., weight importance decreases from red connections to blue connections). The operation of ``scrolling'' reassigns the less important parameters from the previous task to become the most important parameters for the current task, and vice versa for the most important parameters from the previous task. Consider the assignment of weight importance ranking at the first task, as shown in Eq.~\ref{ranking_1}. The weights will be scrolled accordingly with scrolling step size $S$ at task $t$ as
\begin{equation}
\theta^{1} = \theta_{w_{(t\%N)*S}}, \theta^{2} = \theta_{w_{(t\%N)*S}} \cup \theta_{w_{(t\%N)*S+1}}, ..., \theta^{N} = \theta.
\label{ranking_2}
\end{equation}
To implement this updated assignment, the same loss function as in Eq.~\ref{loss_dynamic} will be utilized during training for task $t$.

\subsection{Combination of ScrollNet and continual learning methods}
To demonstrate the orthogonality of our proposed \emph{ScrollNet} to regularization-based and replay-based continual learning methods, we first rephrase Eq.~\ref{loss_dynamic} as
\begin{equation}
\mathcal{L}_{dynamic} = \mathcal{L} \left(f_{N}(\mathcal{X}_{t}; \theta^{N}), \mathcal{Y}_{t}\right) + \sum_{n=1}^{N-1} \mathcal{L}\left(f_{n}(\mathcal{X}_{t}; \theta^{n}), \mathcal{Y}_{t}\right).
\label{loss_dynamic_decouple}
\end{equation}
Then we can see that the first term in the loss function of \emph{ScrollNet} is a normal cross-entropy loss at each task which encompasses the entire network (equivalent to the loss function in Eq.~\ref{cl_loss}). Additionally, it incorporates a series of cross-entropy losses applied to other sub-networks for realizing weight ranking. Here, we present two examples showcasing the combination of ScrollNet with LwF~\cite{li2017learning} and the combination of ScrollNet with EWC~\cite{kirkpatrick2017overcoming}:

\noindent (1) Loss function of \textbf{ScrollNet + LwF}:
\begin{equation}
\begin{split}
\mathcal{L}_{N} \left(f_{N}(\mathcal{X}_{t}; \theta_{new}^{N}), \mathcal{Y}_{t}\right)&+\lambda*\mathcal{L}_{O} \left(f_{N}(\mathcal{X}_{t}; \theta_{new}^{N}), f_{N}(\mathcal{X}_{t}; \theta_{old}^{N})\right) \\
&+ \sum_{n=1}^{N-1} \mathcal{L}_{N}\left(f_{n}(\mathcal{X}_{t}; \theta_{new}^{n}), \mathcal{Y}_{t}\right),
\end{split}
\label{loss_dynamic_lwf}
\end{equation}
where $\mathcal{L}_{N}$ and $\mathcal{L}_{O}$ represent the cross-entropy loss applied to the output of the new task and old task for new data, respectively. $\theta_{new}$ and $\theta_{old}$ denote the parameters of the current model and the frozen old model, respectively. $\lambda$ is a hyperparameter that determines the weight of regularization strength. In our experiments, we set $\lambda$ to 1. 

\noindent (2) Loss function of \textbf{ScrollNet + EWC}:
\begin{equation}
\begin{split}
\mathcal{L}_{N} \left(f_{N}(\mathcal{X}_{t}; \theta_{new}^{N}), \mathcal{Y}_{t}\right)&+\sum_{i}\frac{\lambda}{2}*F^{i}(\theta_{new}^{N,i}-\theta_{old}^{N,i})^2 \\
&+ \sum_{n=1}^{N-1} \mathcal{L}_{N}\left(f_{n}(\mathcal{X}_{t}; \theta_{new}^{n}), \mathcal{Y}_{t}\right),
\end{split}
\label{loss_dynamic_ewc}
\end{equation}
where F is the Fisher information matrix and $i$ labels each parameter. We set $\lambda$ to 5000 in our experiments. 

Please see the pseudo algorithm.~\ref{alg:algorithm1} for the summarized optimizing procedure of our \emph{ScrollNet}.

\begin{algorithm}[ht]
\caption{Scrolling Neural Network (ScrollNet) for CL.}\label{alg:algorithm1}
\renewcommand{\algorithmicrequire}{\textbf{Input:}}  
\begin{algorithmic}[1]
    \Require Sequential data $\{\mathcal{D}_t\}_{t=1}^{\mathcal{T}}$, model weights $\theta$
    \Require Number of sub-networks $N$
    \Require Loss function $\mathcal{L}_{CL}$ of the combined CL method
    \Require Set scrolling step size $S$                \Comment{always 1 in this work}
    \State Randomly initialize $\theta$
    \State Equally split the model as $\theta = \{\theta_{w_{1}}, \theta_{w_{2}}, ..., \theta_{w_{N}}\}$
    \State Initialize sub-networks as Eq.~\ref{ranking_1}
    \For{task $t = 1, . . . , \mathcal{T}$}
        \State Scroll the weights in each sub-network as in Eq.~\ref{ranking_2}
        \For{batch $\mathbf{b}_t\sim\mathcal{D}_t$}
            \State Set Total\_loss = 0
                \For{$n = 1, . . . , N$}
                    \If {$n == N$}
                        \State Calc $\mathcal{L} = \mathcal{L} \left(f_{N}(\mathcal{X}_{t}; \theta^{N}), \mathcal{Y}_{t}\right)+\mathcal{L}_{CL}$
                        \State Total\_loss += $\mathcal{L}$
                    \Else
                        \State Calc $\mathcal{L} = \mathcal{L} \left(f_{n}(\mathcal{X}_{t}; \theta^{n}), \mathcal{Y}_{t}\right)$
                        \State Total\_loss += $\mathcal{L}$
                    \EndIf
                \EndFor
            \State \textit{Backpropagation} with Total\_loss
            \State Update model weights $\theta$
        \EndFor
    \EndFor
\end{algorithmic}
\end{algorithm}

\section{Experiments}

\subsection{Experimental settings}
\noindent \textbf{Datasets.} We evaluate performance on two datasets: CIFAR100~\cite{krizhevsky2009learning}, and TinyImageNet~\cite{le2015tiny}.
CIFAR100 contains 100 classes, each with 600 images, among which 500 images are for training and the other 100 are for test usage.
Tiny ImageNet contains 100,000 images of 200 classes (500 for each class) downsized to 64×64 colored images. Each class has 500 training images, 50 validation images and 50 test images. In our experiments, we consider different numbers of dataset splits (e.g., 5 splits, 10 splits, and 20 splits) to verify the effectiveness of our proposed method on various lengths of sequential tasks.

\begin{figure}[t]
  \centering
   \includegraphics[width=0.95\linewidth]{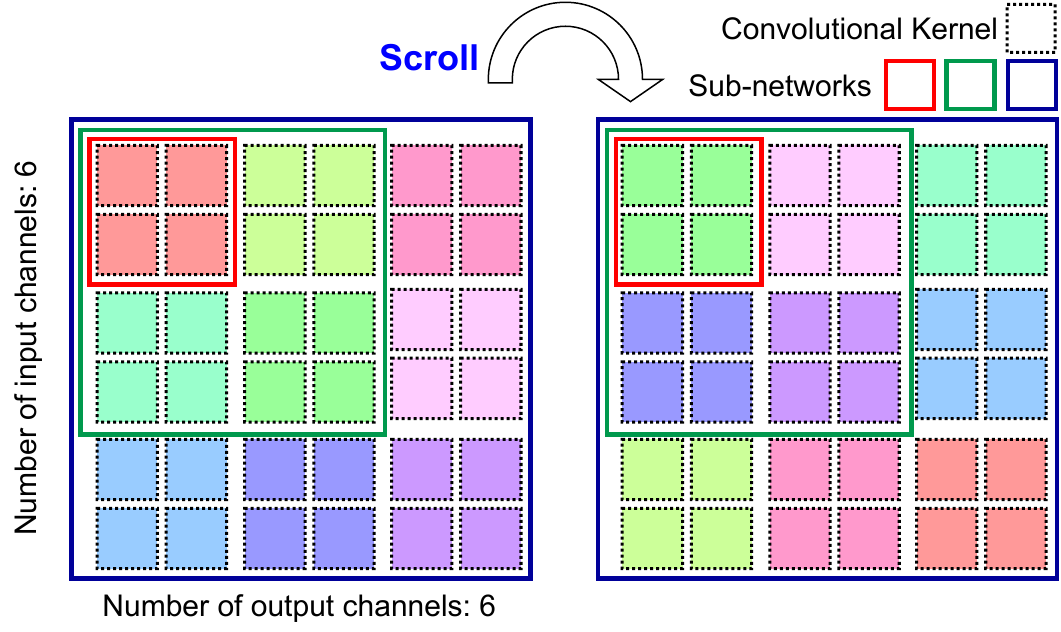}
   \caption{Illustration for one scrolling convolutional layer with three switchable widths as 2, 4, and 6.}
   \label{fig:scrollconv}
   \vspace{-0.3cm}
\end{figure}

\noindent \textbf{Architecture.} In our experiments, our proposed \ourmethod is built on ResNet18~\cite{he2016deep}, a commonly used architecture in the literature for CIFAR100 and TinyImageNet datasets. Firstly, following \cite{yu2018slimmable}, we transform ResNet18 into a channel-wise dynamic network by incorporating slimmable convolutional layers, slimmable FC layers (used as classification heads), and switchable batch normalization layers instead of normal convolutional layers, FC layers, and standard batch normalization layers. In our experiments, we build the slimmable ResNet18 with two different numbers of sub-networks, 2 and 4. Subsequently, we implement \textit{scrolling} for the slimmable convolutional layers and the slimmable FC heads, moving them in a predetermined direction when starting a new task, as depicted in Figure~\ref{fig:scrollconv}. Please see Figure~\ref{ScrollNet} for the \textit{scrolling} FC layer. Note that we do not apply \textit{scrolling} to switchable batch normalization layers since satisfactory performance was observed with a fixed order in sequential tasks. Finally, we implement two variants of our proposed method, namely ScrollNet-2 and ScrollNet-4, which involve 2 and 4 sub-networks (or can be viewed as the number of model splits) in the slimmable ResNet18, respectively. In this paper, we set the step size for scrolling constantly as $S=1$.

\noindent \textbf{Baselines.} Besides the standard fine-tuning, to demonstrate the efficacy of the proposed \ourmethod mechanism, we integrate it with different continual learning approaches. Specifically, we combine \ourmethod with three exemplar-free methods such as EWC~\cite{kirkpatrick2017overcoming}, MAS~\cite{aljundi2018memory}, and LwF~\cite{li2017learning}. 
Additionally, we also incorporate \ourmethod with three exemplar-based frameworks including iCaRL~\cite{rebuffi2017icarl}, LUCIR~\cite{hou2019learning}, and BiC~\cite{wu2019large}. 
We report average accuracy at each task on both \textit{task-aware} and \textit{task-agnostic} settings. Note that the implementation of all these baseline methods is from the CL framework\footnote{\url{https://github.com/mmasana/FACIL}} proposed in \cite{masana2022class}. 

\noindent \textbf{Training details.} We train the model for 200 epochs per task for different numbers of splits. The learning rate is initialized to 0.1 and is decayed by a rate of 0.1 at the 80th and 120th epochs. We use an SGD optimizer with a batch size of 64.
For the exemplar-based methods, we utilize 2000 exemplars selected with herding~\cite{rebuffi2017icarl}, constituting a fixed memory.

\begin{table*}[tb!]
    \centering
    \scalebox{0.99}{
    \begin{tabular}{l|c|ccc|ccc}
    \hline
        \multirow{2}{*}{\textbf{Method}} &\multirow{2}{*}{\textbf{Exemplar}} &\multicolumn{3}{c|}{\textbf{Task-agnostic}} &\multicolumn{3}{c}{\textbf{Task-aware}}\\
        
        & &5 splits &10 splits &20 splits &5 splits &10 splits &20 splits\\
    \hline     
       FT  &No &25.47\mypm{}0.35 &15.17\mypm{}1.06 &7.27\mypm{}0.67 &54.23\mypm{}1.46 &49.97\mypm{}2.35 &42.67\mypm{}4.15\\
       \;\;ScrollNet-2  &No &28.70\mypm{}0.56 &16.60\mypm{}0.26  &8.13\mypm{}0.31 &58.17\mypm{}1.00 &53.00\mypm{}1.56 &49.70\mypm{}1.84\\
       \;\;ScrollNet-4   &No &\textbf{31.67\mypm{}1.59} &\textbf{18.66\mypm{}0.49}  &\textbf{9.33\mypm{}0.95} &\textbf{60.47\mypm{}1.53}
       &\textbf{56.60\mypm{}1.35} &\textbf{55.27\mypm{}2.11}\\
        \hline
       LwF  &No &43.80\mypm{}0.35 &30.46\mypm{}1.74 &17.90\mypm{}0.62 &78.83\mypm{}0.76 &\textbf{80.97\mypm{}0.91} &80.30\mypm{}0.70\\
       \;\;ScrollNet-2   &No &45.67\mypm{}0.67 &30.63\mypm{}2.76  &17.93\mypm{}1.27  &\textbf{79.77\mypm{}0.72} &80.40\mypm{}0.85 &79.47\mypm{}0.90\\
       \;\;ScrollNet-4   &No &\textbf{46.70\mypm{}0.60} &\textbf{32.37\mypm{}0.76}  &\textbf{19.60\mypm{}0.56}  &79.30\mypm{}0.72 &79.93\mypm{}0.96 &\textbf{81.47\mypm{}1.06}\\
         \hline
       EWC  &No &31.57\mypm{}0.72 &19.56\mypm{}0.65 &9.80\mypm{}1.40 &61.43\mypm{}0.64 &57.47\mypm{}2.54 &51.33\mypm{}1.70\\
       \;\;ScrollNet-2   &No &39.47\mypm{}0.76 &26.57\mypm{}1.27 &13.23\mypm{}1.27 &69.03\mypm{}0.40 &67.00\mypm{}2.57 &62.80\mypm{}0.88\\
       \;\;ScrollNet-4   &No &\textbf{40.97\mypm{}0.75} &\textbf{28.03\mypm{}1.88} &\textbf{15.41\mypm{}1.85} &\textbf{71.47\mypm{}0.85} &\textbf{69.87\mypm{}2.15} &\textbf{67.97\mypm{}1.63}\\
         \hline
       MAS  &No &35.23\mypm{}0.49 &21.50\mypm{}1.48 &9.70\mypm{}0.98 &65.03\mypm{}0.21 &61.53\mypm{}2.65 &55.20\mypm{}2.26\\
       \;\;ScrollNet-2   &No &39.83\mypm{}0.90 &25.93\mypm{}2.21 &13.80\mypm{}1.85 &69.03\mypm{}0.31 &68.33\mypm{}2.79 &64.73\mypm{}0.65\\
       \;\;ScrollNet-4   &No &\textbf{40.04\mypm{}1.11} &\textbf{27.40\mypm{}0.52} &\textbf{15.50\mypm{}1.32} &\textbf{70.40\mypm{}1.14} &\textbf{69.83\mypm{}1.06} &\textbf{69.97\mypm{}1.02}\\
        \hline
       iCaRL  &2000 &53.76\mypm{}0.55 &42.83\mypm{}2.38 &32.57\mypm{}1.80 &77.07\mypm{}0.59 &78.97\mypm{}2.15 &81.10\mypm{}1.39\\
       \;\;ScrollNet-2   &2000 &\textbf{54.90\mypm{}0.46} &\textbf{45.30\mypm{}2.21} &35.93\mypm{}1.60 &77.87\mypm{}0.61 &80.73\mypm{}1.75 &82.97\mypm{}1.01\\
       \;\;ScrollNet-4   &2000 &54.80\mypm{}0.74 &44.87\mypm{}1.17 &\textbf{37.03\mypm{}1.53} &\textbf{77.93\mypm{}0.35} &\textbf{81.17\mypm{}1.04} &\textbf{83.40\mypm{}1.56}\\
         \hline
       BiC  &2000 &58.20\mypm{}0.62 &49.26\mypm{}1.05 &37.70\mypm{}1.51 &80.60\mypm{}0.26 &83.43\mypm{}1.17 &85.40\mypm{}0.62\\
       \;\;ScrollNet-2   &2000 &58.50\mypm{}0.36 &49.67\mypm{}1.44 &\textbf{39.00\mypm{}1.44} &\textbf{81.27\mypm{}0.42} &\textbf{83.60\mypm{}1.45} &\textbf{85.97\mypm{}0.68}\\
       \;\;ScrollNet-4   &2000 &\textbf{58.77\mypm{}0.87} &\textbf{49.72\mypm{}1.73} &38.64\mypm{}0.47 &81.00\mypm{}0.47 &83.47\mypm{}1.50 &85.73\mypm{}0.57\\
         \hline
       LUCIR  &2000 &54.80\mypm{}0.82 &41.97\mypm{}1.80 &34.23\mypm{}0.51 &81.03\mypm{}0.15 &83.23\mypm{}1.50 &85.37\mypm{}0.76\\
       \;\;ScrollNet-2   &2000 &54.50\mypm{}0.78 &42.90\mypm{}1.41 &34.83\mypm{}1.76 &80.87\mypm{}0.57 &83.43\mypm{}1.29 &85.17\mypm{}1.07\\
       \;\;ScrollNet-4   &2000 &\textbf{55.53\mypm{}0.82} &\textbf{45.46\mypm{}0.58 } &\textbf{37.23\mypm{}1.00} &\textbf{81.60\mypm{}0.69} &\textbf{84.13\mypm{}1.10} &\textbf{86.10\mypm{}1.05}\\
    \hline
    \end{tabular}
    }
    \vspace{0.3cm}
    \caption{Average accuracy after the last task for various continual learning methods and their combination with ScrollNet-N (N is the number of sub-networks). We run experiments three times with random class orders on CIFAR100 and report averages\mypm{}deviations.}
    \label{tab:cifar}
\end{table*}

\begin{table*}[tb!]
    \centering
    \scalebox{0.99}{
    \begin{tabular}{l|c|ccc|ccc}
    \hline
        \multirow{2}{*}{\textbf{Method}} &\multirow{2}{*}{\textbf{Exemplar}} &\multicolumn{3}{c|}{\textbf{Task-agnostic}} &\multicolumn{3}{c}{\textbf{Task-aware}}\\
        
        & &5 splits &10 splits &20 splits &5 splits &10 splits &20 splits\\
    \hline
       FT  &No &18.60\mypm{}0.46 &10.70\mypm{}0.30 &5.80\mypm{}0.62 &37.20\mypm{}1.13 &31.93\mypm{}1.62 &28.67\mypm{}1.18\\
       \;\;ScrollNet-2   &No &\textbf{20.77\mypm{}1.21} &\textbf{12.60\mypm{}0.61}  &6.97\mypm{}0.68 &41.83\mypm{}1.50 &38.67\mypm{}1.25 &35.83\mypm{}1.50\\
       \;\;ScrollNet-4  &No &20.50\mypm{}0.56 &12.37\mypm{}0.61  &\textbf{7.63\mypm{}0.81} &\textbf{43.60\mypm{}0.75} &\textbf{40.03\mypm{}0.70} &\textbf{40.07\mypm{}1.47}\\
    \hline
       LwF  &No &34.67\mypm{}0.76 &24.23\mypm{}1.66 &15.83\mypm{}1.22 &65.87\mypm{}0.29 &67.93\mypm{}1.10 &68.03\mypm{}1.12\\
       \;\;ScrollNet-2   &No &35.76\mypm{}0.84 &24.80\mypm{}1.74  &15.33\mypm{}1.38  &\textbf{66.40\mypm{}0.61} &\textbf{68.03\mypm{}0.90} &67.27\mypm{}1.23\\
       \;\;ScrollNet-4   &No &\textbf{36.47\mypm{}0.93} &\textbf{24.88\mypm{}1.17}  &\textbf{18.33\mypm{}1.40}  &65.70\mypm{}0.00 &66.97\mypm{}1.12 &\textbf{72.13\mypm{}0.56}\\
         \hline  
       EWC  &No &29.37\mypm{}0.90 &19.37\mypm{}1.08 &9.87\mypm{}0.42 &54.03\mypm{}1.70 &51.07\mypm{}0.38 &44.27\mypm{}1.21\\
       \;\;ScrollNet-2   &No &\textbf{32.00\mypm{}0.10} &24.70\mypm{}0.98 &15.17\mypm{}0.21 &\textbf{60.73\mypm{}0.83} &61.27\mypm{}0.64 &60.40\mypm{}2.10\\
       \;\;ScrollNet-4   &No &30.87\mypm{}1.46 &\textbf{25.23\mypm{}0.38} &\textbf{17.23\mypm{}1.02} &59.70\mypm{}2.04 &\textbf{64.23\mypm{}0.40} &\textbf{64.00\mypm{}1.68}\\
         \hline  
       MAS  &No &27.57\mypm{}0.90 &16.43\mypm{}0.64 &9.40\mypm{}0.35 &51.57\mypm{}1.51 &47.70\mypm{}0.46 &45.40\mypm{}1.90\\
       \;\;ScrollNet-2   &No &\textbf{31.70\mypm{}0.36} &21.70\mypm{}0.62 &12.77\mypm{}0.42 &58.03\mypm{}0.67 &57.57\mypm{}0.60 &55.63\mypm{}1.63\\
       \;\;ScrollNet-4   &No &30.57\mypm{}1.40 &\textbf{24.30\mypm{}0.56} &\textbf{15.63\mypm{}0.86} &\textbf{58.30\mypm{}2.00} &\textbf{60.67\mypm{}0.42} &\textbf{62.03\mypm{}1.89}\\        
    \hline
       iCaRL  &2000 & 37.10\mypm{}0.70 &31.50\mypm{}0.78 &21.13\mypm{}0.68 &60.57\mypm{}0.93 &67.50\mypm{}1.21 &68.57\mypm{}1.01\\
       \;\;ScrollNet-2   &2000 &\textbf{39.40\mypm{}0.87} &32.6\mypm{}1.06 &23.17\mypm{}1.45 &\textbf{61.80\mypm{}0.70} &68.37\mypm{}1.46 &70.3\mypm{}1.65\\
       \;\;ScrollNet-4   &2000 &38.93\mypm{}0.91 &\textbf{33.97\mypm{}1.11} &\textbf{25.33\mypm{}1.24} &61.63\mypm{}0.67 &\textbf{69.10\mypm{}1.04} &\textbf{71.17\mypm{}1.10}\\
         \hline  
       BiC  &2000 &43.70\mypm{}0.72 &36.03\mypm{}0.50 &26.03\mypm{}1.65 &66.73\mypm{}0.92 &71.80\mypm{}1.13 &\textbf{76.03\mypm{}1.12}\\
       \;\;ScrollNet-2   &2000 &45.00\mypm{}0.40 &36.93\mypm{}0.55 &26.87\mypm{}1.10 &\textbf{68.17\mypm{}0.50} &72.27\mypm{}0.83 &75.73\mypm{}0.66\\
       \;\;ScrollNet-4   &2000 &\textbf{45.10\mypm{}1.05} &\textbf{37.17\mypm{}0.29} &\textbf{27.30\mypm{}1.21} &67.93\mypm{}0.15 &\textbf{72.33\mypm{}0.96} &75.50\mypm{}0.82\\
         \hline  
       LUCIR  &2000 &\textbf{36.97\mypm{}1.06} &24.67\mypm{}1.00 &17.73\mypm{}1.46 &\textbf{66.70\mypm{}0.92} &\textbf{70.30\mypm{}0.44} &73.33\mypm{}0.40\\
       \;\;ScrollNet-2   &2000 &35.23\mypm{}0.29 &23.90\mypm{}1.71 &16.83\mypm{}1.25 &65.53\mypm{}0.78 &68.93\mypm{}1.50 &72.33\mypm{}0.51\\
       \;\;ScrollNet-4   &2000 &34.43\mypm{}0.67 &\textbf{24.77\mypm{}1.50} &\textbf{18.17\mypm{}0.81} &66.17\mypm{}0.60 &70.27\mypm{}0.85 &\textbf{74.53\mypm{}0.60}\\
    \hline
    \end{tabular}
    }
    \vspace{0.3cm}
    \caption{Average accuracy after the last task for various continual learning methods and their combination with ScrollNet-N (N is the number of sub-networks). We run experiments three times with random class orders on TinyImageNet and report averages\mypm{}deviations.}
    \label{tab:tiny}
\end{table*}

\begin{figure*}[t]
  \centering
   \includegraphics[width=1\linewidth]{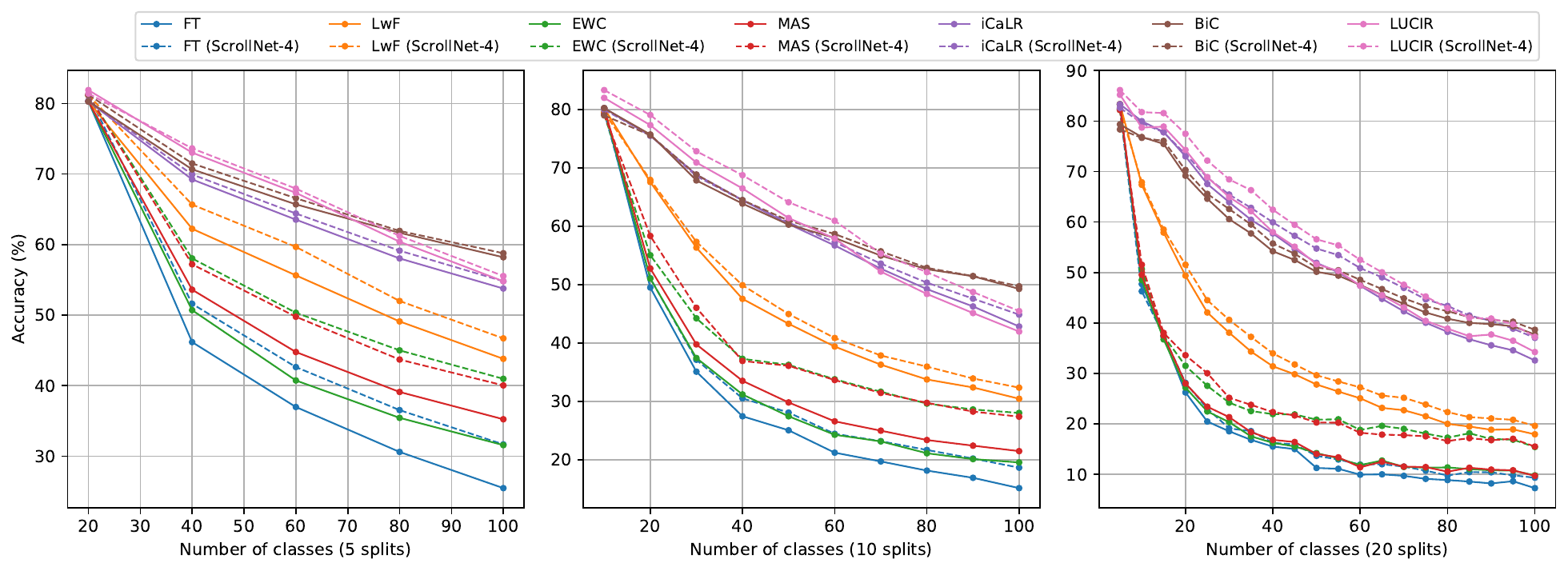}
   \caption{Results on CIFAR100 dataset (\textbf{task-agnostic}). We report the average accuracies of three runs after each task, each with a random class order.}
   \label{CIFAR_tag}
   \vspace{-0.2cm}
\end{figure*}

\begin{figure*}[t]
  \centering
   \includegraphics[width=1\linewidth]{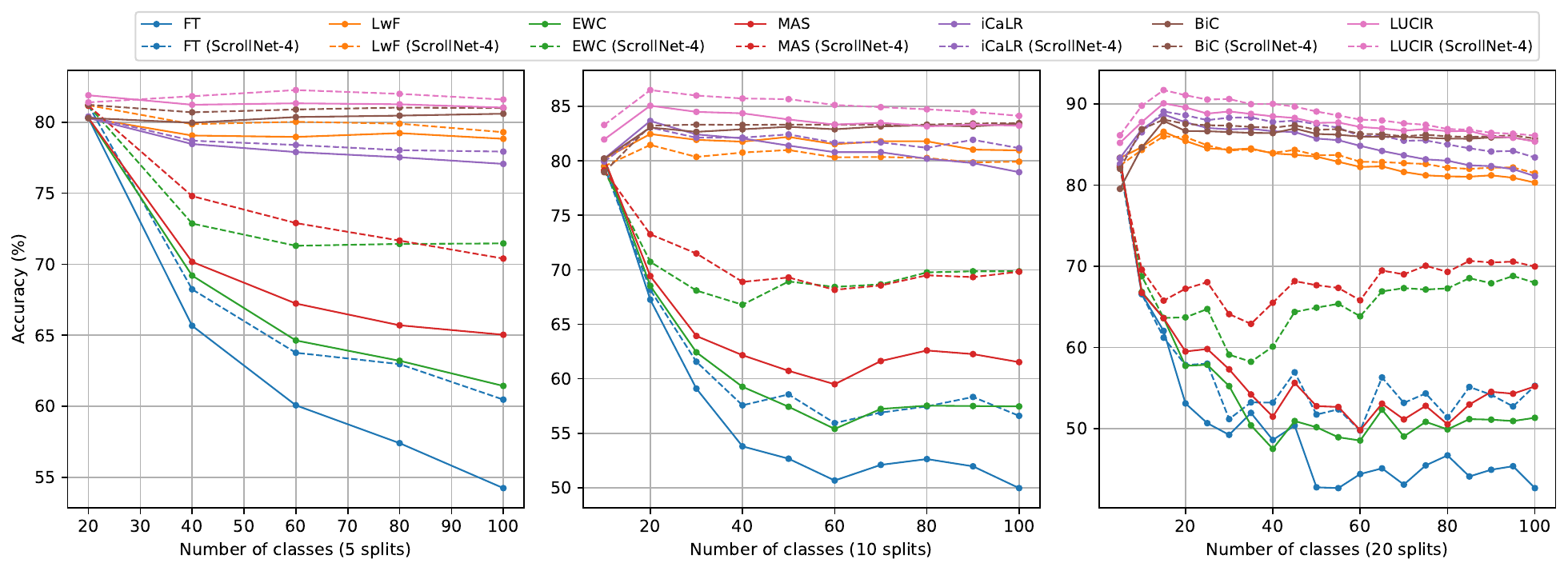}
   \caption{Results on CIFAR100 dataset (\textbf{task-aware}). We report the average accuracies of three runs after each task, each with a random class order.}
   \label{CIFAR_taw}
   \vspace{-0.2cm}
\end{figure*}

\subsection{Main results}
In Table~\ref{tab:cifar} and Table~\ref{tab:tiny}, we present the average accuracy after the last task for all baseline methods, as well as their combinations with our proposed \ourmethod, on CIFAR100 and TinyImageNet, respectively. Additionally, Figure~\ref{CIFAR_tag} and Figure~\ref{CIFAR_taw} display the curves of average accuracies at each task on CIFAR100, representing the task-agnostic and task-aware settings, respectively.

From the tables, we observe that \ourmethod, when combined with various CL methods, outperforms most of their corresponding baselines in the task-agnostic setting across all different dataset splits, except for only LUCIR on TinyImageNet with 5 splits. Notably, \ourmethod demonstrates particularly promising results when paired with parameter regularization-based CL methods, such as EWC and MAS. For instance, when combined with EWC, the highest improvements reach up to 9.40\% (5 splits), 8.47\% (10 splits), and 5.61\% (20 splits) on CIFAR100, as well as 2.67\% (5 splits), 5.86\% (10 splits), and 5.36\%(20 splits) on TinyImageNet. It is worth mentioning that while EWC was not originally proposed for class incremental learning, the improvements observed in the task incremental learning setting (task-aware) are even higher, reaching 10.04\% (5 splits), 12.40\% (10 splits), and 16.64\% (20 splits) on CIFAR100, and 6.70\% (5 splits), 13.16\% (10 splits), and 19.73\% on TinyImageNet. Those results verify the orthogonality of our proposed method to regularization-based and replay-based CL methods, especially to the parameter regularization-based methods. Furthermore, we observe that the performance improvements brought by \emph{ScrollNet} in the task-agnostic setting are comparatively less pronounced on TinyImageNet than on CIFAR100. This could be attributed to the limited capacity of ResNet18 for handling larger datasets, thereby constraining the effectiveness of \ourmethod.

Regarding the task-aware setting, our method also outperforms most of the baselines, except for LwF on CIFAR100 with 10 splits, BiC on TinyImageNet with 20 splits, and LUCIR on TinyImageNet with 5/10 splits. One possible reason for this is that the accuracies of these baselines in the task-aware setting are already quite high, leaving limited room for further improvement. Additionally, we find that the benefits of combining \ourmethod and LwF are not evident in other cases within the task-aware setting. This suggests that the advantage of incorporating \ourmethod with LwF in terms of cross-task representation learning may diminish in the task-aware setting. As for BiC and LUCIR, similar to the previous discussion, the limited capacity of ResNet18 on TinyImageNet also constrains the performance of our method.

Comparing the performance of the two variants of \emph{ScrollNet}, we find that ScrollNet-4 consistently outperforms ScrollNet-2 in most cases. The reason behind this is that more model splitting corresponds to a finer weight ranking, providing a more precise ``scrolling'' mechanism that achieves a better stability-plasticity tradeoff. This holds particularly true for longer sequences of tasks, such as 20 splits, where \ourmethod with more model splitting yields superior results.
\section{Discussion and future works}
We believe that our work presents a novel direction for the continual learning community, as it introduces the pre-assignment of weight importance ranking before learning each task using a dynamic network, and provides an explicit way to strike a balance between stability and plasticity. The experiments have demonstrated the effectiveness of our proposed method. However, it is important to note that in this work, we have just implemented a simple dynamic network, i.e., considering only channel-wise model splitting and up to four splits throughout the network. Furthermore, our strategy (i.e., ``scrolling'') to leverage weight ranking knowledge for improving the performance of various CL methods is straightforward. Moving forward, we plan to explore more refined splitting techniques borrowing from the field of dynamic networks, and investigate alternative strategies for utilizing this prior knowledge. For example, one potential approach could involve calculating the step size of `scrolling' based on the correlation between each task. Given that the performance improvement on TinyImageNet is not particularly evident, we are also planning to explore the effectiveness of our method when using larger models, such as ResNet101 and ViTs~\cite{dosovitskiy2020image}.

\section{Conclusion}
In this paper, we introduce ScrollNet, a scrolling neural network designed for continual learning. ScrollNet functions as a dynamic network that assigns the ranking of weight importance for each task before data exposure, thereby achieving a more favorable tradeoff between stability and plasticity during sequential task learning by adjusting this ranking for different tasks. 
Furthermore, we demonstrate that ScrollNet can be combined with various continual learning methods, including regularization-based and replay-based approaches. 
We validate the effectiveness of our proposed method through experiments conducted on CIFAR100 and TinyImagenet datasets. 

\section{Limitation}
The training time will increase when combined with our proposed \emph{ScrollNet} due to multiple forward passes with different sub-networks during training. This is especially true when \emph{ScrollNet} has a larger number of model splittings. However, during inference, the computational cost of combining \emph{ScrollNet} will remain the same because only the whole model will be executed.

\section*{Acknowledgements}
We acknowledge the support from the Spanish Government funding for projects PID2022-143257NB-I00, TED2021-132513B-I00.

{\small
\bibliographystyle{ieee_fullname}
\bibliography{egbib}

\begin{thebibliography}{10}\itemsep=-1pt

\bibitem{abati2020conditional}
Davide Abati, Jakub Tomczak, Tijmen Blankevoort, Simone Calderara, Rita
  Cucchiara, and Babak~Ehteshami Bejnordi.
\newblock Conditional channel gated networks for task-aware continual learning.
\newblock In {\em Proceedings of the IEEE/CVF Conference on Computer Vision and
  Pattern Recognition}, pages 3931--3940, 2020.

\bibitem{aljundi2018memory}
Rahaf Aljundi, Francesca Babiloni, Mohamed Elhoseiny, Marcus Rohrbach, and
  Tinne Tuytelaars.
\newblock Memory aware synapses: Learning what (not) to forget.
\newblock In {\em ECCV}, pages 139--154, 2018.

\bibitem{bengio2013estimating}
Yoshua Bengio, Nicholas L{\'e}onard, and Aaron Courville.
\newblock Estimating or propagating gradients through stochastic neurons for
  conditional computation.
\newblock {\em arXiv preprint arXiv:1308.3432}, 2013.

\bibitem{bolukbasi2017adaptive}
Tolga Bolukbasi, Joseph Wang, Ofer Dekel, and Venkatesh Saligrama.
\newblock Adaptive neural networks for efficient inference.
\newblock In {\em International Conference on Machine Learning}, pages
  527--536. PMLR, 2017.

\bibitem{buzzega2020dark}
Pietro Buzzega, Matteo Boschini, Angelo Porrello, Davide Abati, and Simone
  Calderara.
\newblock Dark experience for general continual learning: a strong, simple
  baseline.
\newblock {\em Advances in neural information processing systems},
  33:15920--15930, 2020.

\bibitem{chaudhry2018efficient}
Arslan Chaudhry, Marc’Aurelio Ranzato, Marcus Rohrbach, and Mohamed
  Elhoseiny.
\newblock Efficient lifelong learning with a-gem.
\newblock In {\em International Conference on Learning Representations}, 2018.

\bibitem{chaudhry2019tiny}
Arslan Chaudhry, Marcus Rohrbach, Mohamed Elhoseiny, Thalaiyasingam Ajanthan,
  Puneet~K Dokania, Philip~HS Torr, and Marc'Aurelio Ranzato.
\newblock On tiny episodic memories in continual learning.
\newblock {\em arXiv preprint arXiv:1902.10486}, 2019.

\bibitem{cho2014exponentially}
Kyunghyun Cho and Yoshua Bengio.
\newblock Exponentially increasing the capacity-to-computation ratio for
  conditional computation in deep learning.
\newblock {\em arXiv preprint arXiv:1406.7362}, 2014.

\bibitem{de2019continual}
Matthias De~Lange, Rahaf Aljundi, Marc Masana, Sarah Parisot, Xu Jia, Ales
  Leonardis, Gregory Slabaugh, and Tinne Tuytelaars.
\newblock A continual learning survey: Defying forgetting in classification
  tasks.
\newblock {\em IEEE TPAMI}, 2021.

\bibitem{delange2021continual}
Matthias Delange, Rahaf Aljundi, Marc Masana, Sarah Parisot, Xu Jia, Ales
  Leonardis, Greg Slabaugh, and Tinne Tuytelaars.
\newblock A continual learning survey: Defying forgetting in classification
  tasks.
\newblock {\em IEEE TPAMI}, 2021.

\bibitem{dosovitskiy2020image}
Alexey Dosovitskiy, Lucas Beyer, Alexander Kolesnikov, Dirk Weissenborn,
  Xiaohua Zhai, Thomas Unterthiner, Mostafa Dehghani, Matthias Minderer, Georg
  Heigold, Sylvain Gelly, et~al.
\newblock An image is worth 16x16 words: Transformers for image recognition at
  scale.
\newblock {\em arXiv preprint arXiv:2010.11929}, 2020.

\bibitem{douillard2021plop}
Arthur Douillard, Yifu Chen, Arnaud Dapogny, and Matthieu Cord.
\newblock Plop: Learning without forgetting for continual semantic
  segmentation.
\newblock In {\em Proceedings of the IEEE/CVF conference on computer vision and
  pattern recognition}, pages 4040--4050, 2021.

\bibitem{eigen2013learning}
David Eigen, Marc'Aurelio Ranzato, and Ilya Sutskever.
\newblock Learning factored representations in a deep mixture of experts.
\newblock {\em arXiv preprint arXiv:1312.4314}, 2013.

\bibitem{frosst2017distilling}
Nicholas Frosst and Geoffrey Hinton.
\newblock Distilling a neural network into a soft decision tree.
\newblock {\em arXiv preprint arXiv:1711.09784}, 2017.

\bibitem{graves2016adaptive}
Alex Graves.
\newblock Adaptive computation time for recurrent neural networks.
\newblock {\em arXiv preprint arXiv:1603.08983}, 2016.

\bibitem{han2021dynamic}
Yizeng Han, Gao Huang, Shiji Song, Le Yang, Honghui Wang, and Yulin Wang.
\newblock Dynamic neural networks: A survey.
\newblock {\em IEEE Transactions on Pattern Analysis and Machine Intelligence},
  44(11):7436--7456, 2021.

\bibitem{hayes2020remind}
Tyler~L Hayes, Kushal Kafle, Robik Shrestha, Manoj Acharya, and Christopher
  Kanan.
\newblock Remind your neural network to prevent catastrophic forgetting.
\newblock In {\em ECCV}, pages 466--483. Springer, 2020.

\bibitem{he2016deep}
Kaiming He, Xiangyu Zhang, Shaoqing Ren, and Jian Sun.
\newblock Deep residual learning for image recognition.
\newblock In {\em Proceedings of the IEEE conference on computer vision and
  pattern recognition}, pages 770--778, 2016.

\bibitem{hou2019learning}
Saihui Hou, Xinyu Pan, Chen~Change Loy, Zilei Wang, and Dahua Lin.
\newblock Learning a unified classifier incrementally via rebalancing.
\newblock In {\em CVPR}, pages 831--839, 2019.

\bibitem{huang2017multi}
Gao Huang, Danlu Chen, Tianhong Li, Felix Wu, Laurens Van Der~Maaten, and
  Kilian~Q Weinberger.
\newblock Multi-scale dense networks for resource efficient image
  classification.
\newblock {\em arXiv preprint arXiv:1703.09844}, 2017.

\bibitem{jacobs1991adaptive}
Robert~A Jacobs, Michael~I Jordan, Steven~J Nowlan, and Geoffrey~E Hinton.
\newblock Adaptive mixtures of local experts.
\newblock {\em Neural computation}, 3(1):79--87, 1991.

\bibitem{kaya2019shallow}
Yigitcan Kaya, Sanghyun Hong, and Tudor Dumitras.
\newblock Shallow-deep networks: Understanding and mitigating network
  overthinking.
\newblock In {\em International conference on machine learning}, pages
  3301--3310. PMLR, 2019.

\bibitem{kirkpatrick2017overcoming}
James Kirkpatrick, Razvan Pascanu, Neil Rabinowitz, Joel Veness, Guillaume
  Desjardins, Andrei~A Rusu, Kieran Milan, John Quan, Tiago Ramalho, Agnieszka
  Grabska-Barwinska, et~al.
\newblock Overcoming catastrophic forgetting in neural networks.
\newblock {\em Proceedings of the national academy of sciences},
  114(13):3521--3526, 2017.

\bibitem{kong2022balancing}
Yajing Kong, Liu Liu, Zhen Wang, and Dacheng Tao.
\newblock Balancing stability and plasticity through advanced null space in
  continual learning.
\newblock In {\em European Conference on Computer Vision}, pages 219--236.
  Springer, 2022.

\bibitem{kontschieder2015deep}
Peter Kontschieder, Madalina Fiterau, Antonio Criminisi, and Samuel~Rota Bulo.
\newblock Deep neural decision forests.
\newblock In {\em Proceedings of the IEEE international conference on computer
  vision}, pages 1467--1475, 2015.

\bibitem{krizhevsky2009learning}
Alex Krizhevsky.
\newblock Learning multiple layers of features from tiny images.
\newblock 2009.

\bibitem{le2015tiny}
Ya Le and Xuan Yang.
\newblock Tiny imagenet visual recognition challenge.
\newblock {\em CS 231N}, 7(7):3, 2015.

\bibitem{li2017learning}
Zhizhong Li and Derek Hoiem.
\newblock Learning without forgetting.
\newblock {\em IEEE transactions on pattern analysis and machine intelligence},
  40(12):2935--2947, 2017.

\bibitem{liu2018dynamic}
Lanlan Liu and Jia Deng.
\newblock Dynamic deep neural networks: Optimizing accuracy-efficiency
  trade-offs by selective execution.
\newblock In {\em Proceedings of the AAAI Conference on Artificial
  Intelligence}, volume~32, 2018.

\bibitem{liu2020generative}
Xialei Liu, Chenshen Wu, Mikel Menta, Luis Herranz, Bogdan Raducanu, Andrew~D
  Bagdanov, Shangling Jui, and Joost van~de Weijer.
\newblock Generative feature replay for class-incremental learning.
\newblock In {\em CVPR}, pages 226--227, 2020.

\bibitem{lopez2017gradient}
David Lopez-Paz and Marc'Aurelio Ranzato.
\newblock Gradient episodic memory for continual learning.
\newblock {\em Advances in neural information processing systems}, 30, 2017.

\bibitem{mallya2018piggyback}
Arun Mallya, Dillon Davis, and Svetlana Lazebnik.
\newblock Piggyback: Adapting a single network to multiple tasks by learning to
  mask weights.
\newblock In {\em Proceedings of the European Conference on Computer Vision
  (ECCV)}, pages 67--82, 2018.

\bibitem{mallya2018packnet}
Arun Mallya and Svetlana Lazebnik.
\newblock Packnet: Adding multiple tasks to a single network by iterative
  pruning.
\newblock In {\em Proceedings of the IEEE conference on Computer Vision and
  Pattern Recognition}, pages 7765--7773, 2018.

\bibitem{masana2022class}
Marc Masana, Xialei Liu, Bart{\l}omiej Twardowski, Mikel Menta, Andrew~D
  Bagdanov, and Joost Van De~Weijer.
\newblock Class-incremental learning: survey and performance evaluation on
  image classification.
\newblock {\em IEEE Transactions on Pattern Analysis and Machine Intelligence},
  45(5):5513--5533, 2022.

\bibitem{masana2020ternary}
Marc Masana, Tinne Tuytelaars, and Joost van~de Weijer.
\newblock Ternary feature masks: continual learning without any forgetting.
\newblock {\em 2nd CLVISION workshop in CVPR 2021}, 2020.

\bibitem{masse2018alleviating}
Nicolas~Y Masse, Gregory~D Grant, and David~J Freedman.
\newblock Alleviating catastrophic forgetting using context-dependent gating
  and synaptic stabilization.
\newblock {\em Proceedings of the National Academy of Sciences},
  115(44):E10467--E10475, 2018.

\bibitem{mccloskey1989catastrophic}
Michael McCloskey and Neal~J Cohen.
\newblock Catastrophic interference in connectionist networks: The sequential
  learning problem.
\newblock In {\em Psychology of learning and motivation}, volume~24, pages
  109--165. Elsevier, 1989.

\bibitem{mehta2021continual}
Nikhil Mehta, Kevin Liang, Vinay~Kumar Verma, and Lawrence Carin.
\newblock Continual learning using a bayesian nonparametric dictionary of
  weight factors.
\newblock In {\em International Conference on Artificial Intelligence and
  Statistics}, pages 100--108. PMLR, 2021.

\bibitem{mirzadeh2020understanding}
Seyed~Iman Mirzadeh, Mehrdad Farajtabar, Razvan Pascanu, and Hassan
  Ghasemzadeh.
\newblock Understanding the role of training regimes in continual learning.
\newblock {\em Advances in Neural Information Processing Systems},
  33:7308--7320, 2020.

\bibitem{odena2017changing}
Augustus Odena, Dieterich Lawson, and Christopher Olah.
\newblock Changing model behavior at test-time using reinforcement learning.
\newblock {\em arXiv preprint arXiv:1702.07780}, 2017.

\bibitem{parisi2019continual}
German~I Parisi, Ronald Kemker, Jose~L Part, Christopher Kanan, and Stefan
  Wermter.
\newblock Continual lifelong learning with neural networks: A review.
\newblock {\em Neural Networks}, 113:54--71, 2019.

\bibitem{rebuffi2017icarl}
Sylvestre-Alvise Rebuffi, Alexander Kolesnikov, Georg Sperl, and Christoph~H
  Lampert.
\newblock icarl: Incremental classifier and representation learning.
\newblock In {\em Proceedings of the IEEE conference on Computer Vision and
  Pattern Recognition}, pages 2001--2010, 2017.

\bibitem{riemer2018learning}
Matthew Riemer, Ignacio Cases, Robert Ajemian, Miao Liu, Irina Rish, Yuhai Tu,
  and Gerald Tesauro.
\newblock Learning to learn without forgetting by maximizing transfer and
  minimizing interference.
\newblock In {\em International Conference on Learning Representations}, 2018.

\bibitem{rusu2016progressive}
Andrei~A Rusu, Neil~C Rabinowitz, Guillaume Desjardins, Hubert Soyer, James
  Kirkpatrick, Koray Kavukcuoglu, Razvan Pascanu, and Raia Hadsell.
\newblock Progressive neural networks.
\newblock {\em arXiv preprint arXiv:1606.04671}, 2016.

\bibitem{serra2018overcoming}
Joan Serra, Didac Suris, Marius Miron, and Alexandros Karatzoglou.
\newblock Overcoming catastrophic forgetting with hard attention to the task.
\newblock In {\em International Conference on Machine Learning}, pages
  4548--4557. PMLR, 2018.

\bibitem{shin2017continual}
Hanul Shin, Jung~Kwon Lee, Jaehong Kim, and Jiwon Kim.
\newblock Continual learning with deep generative replay.
\newblock In {\em NeurIPS}, pages 2994--3003, 2017.

\bibitem{tang2021layerwise}
Shixiang Tang, Dapeng Chen, Jinguo Zhu, Shijie Yu, and Wanli Ouyang.
\newblock Layerwise optimization by gradient decomposition for continual
  learning.
\newblock In {\em Proceedings of the IEEE/CVF conference on Computer Vision and
  Pattern Recognition}, pages 9634--9643, 2021.

\bibitem{teerapittayanon2016branchynet}
Surat Teerapittayanon, Bradley McDanel, and Hsiang-Tsung Kung.
\newblock Branchynet: Fast inference via early exiting from deep neural
  networks.
\newblock In {\em 2016 23rd international conference on pattern recognition
  (ICPR)}, pages 2464--2469. IEEE, 2016.

\bibitem{veit2018convolutional}
Andreas Veit and Serge Belongie.
\newblock Convolutional networks with adaptive inference graphs.
\newblock In {\em Proceedings of the European Conference on Computer Vision
  (ECCV)}, pages 3--18, 2018.

\bibitem{verwimp2023clad}
Eli Verwimp, Kuo Yang, Sarah Parisot, Lanqing Hong, Steven McDonagh, Eduardo
  P{\'e}rez-Pellitero, Matthias De~Lange, and Tinne Tuytelaars.
\newblock Clad: A realistic continual learning benchmark for autonomous
  driving.
\newblock {\em Neural Networks}, 161:659--669, 2023.

\bibitem{von2019continual}
Johannes Von~Oswald, Christian Henning, Benjamin~F Grewe, and Jo{\~a}o
  Sacramento.
\newblock Continual learning with hypernetworks.
\newblock {\em arXiv preprint arXiv:1906.00695}, 2019.

\bibitem{wang2021training}
Shipeng Wang, Xiaorong Li, Jian Sun, and Zongben Xu.
\newblock Training networks in null space of feature covariance for continual
  learning.
\newblock In {\em Proceedings of the IEEE/CVF conference on Computer Vision and
  Pattern Recognition}, pages 184--193, 2021.

\bibitem{wang2018skipnet}
Xin Wang, Fisher Yu, Zi-Yi Dou, Trevor Darrell, and Joseph~E Gonzalez.
\newblock Skipnet: Learning dynamic routing in convolutional networks.
\newblock In {\em Proceedings of the European Conference on Computer Vision
  (ECCV)}, pages 409--424, 2018.

\bibitem{wolczyk2021zero}
Maciej Wo{\l}czyk, Bartosz W{\'o}jcik, Klaudia Ba{\l}azy, Igor~T Podolak, Jacek
  Tabor, Marek {\'S}mieja, and Tomasz Trzcinski.
\newblock Zero time waste: Recycling predictions in early exit neural networks.
\newblock {\em Advances in Neural Information Processing Systems},
  34:2516--2528, 2021.

\bibitem{wortsman2020supermasks}
Mitchell Wortsman, Vivek Ramanujan, Rosanne Liu, Aniruddha Kembhavi, Mohammad
  Rastegari, Jason Yosinski, and Ali Farhadi.
\newblock Supermasks in superposition.
\newblock {\em Advances in Neural Information Processing Systems},
  33:15173--15184, 2020.

\bibitem{wu2018memory}
Chenshen Wu, Luis Herranz, Xialei Liu, Joost van~de Weijer, Bogdan Raducanu,
  et~al.
\newblock Memory replay gans: Learning to generate new categories without
  forgetting.
\newblock {\em NeurIPS}, 31:5962--5972, 2018.

\bibitem{wu2019large}
Yue Wu, Yinpeng Chen, Lijuan Wang, Yuancheng Ye, Zicheng Liu, Yandong Guo, and
  Yun Fu.
\newblock Large scale incremental learning.
\newblock In {\em CVPR}, pages 374--382, 2019.

\bibitem{yang2021slimmable}
Fei Yang, Luis Herranz, Yongmei Cheng, and Mikhail~G Mozerov.
\newblock Slimmable compressive autoencoders for practical neural image
  compression.
\newblock In {\em Proceedings of the IEEE/CVF Conference on Computer Vision and
  Pattern Recognition}, pages 4998--5007, 2021.

\bibitem{yu2018slimmable}
Jiahui Yu, Linjie Yang, Ning Xu, Jianchao Yang, and Thomas Huang.
\newblock Slimmable neural networks.
\newblock In {\em International Conference on Learning Representations}, 2018.

\bibitem{yu2022self}
Lu Yu, Xialei Liu, and Joost Van~de Weijer.
\newblock Self-training for class-incremental semantic segmentation.
\newblock {\em IEEE Transactions on Neural Networks and Learning Systems},
  2022.

\bibitem{zenke2017continual}
Friedemann Zenke, Ben Poole, and Surya Ganguli.
\newblock Continual learning through synaptic intelligence.
\newblock pages 3987--3995. PMLR, 2017.

\end{thebibliography}
}

\end{document}